%% file: main.tex
\def\year{2021}\relax
\documentclass[letterpaper]{article} % DO NOT CHANGE THIS
\usepackage{aaai21}  % DO NOT CHANGE THIS
\usepackage{times}  % DO NOT CHANGE THIS
\usepackage{helvet} % DO NOT CHANGE THIS
\usepackage{courier}  % DO NOT CHANGE THIS
\usepackage[hyphens]{url}  % DO NOT CHANGE THIS
\usepackage{graphicx} % DO NOT CHANGE THIS

\usepackage{scalerel} % scale sub- and upscripts
\usepackage{enumitem}
\usepackage{bbm}
\usepackage{multirow}
\usepackage{subcaption}
\usepackage{booktabs}
\usepackage{amsmath}
\usepackage{tabularx}
\usepackage{ragged2e}
\usepackage{amsfonts}

\usepackage{natbib}  % DO NOT CHANGE THIS AND DO NOT ADD ANY OPTIONS TO IT
\usepackage{caption} % DO NOT CHANGE THIS AND DO NOT ADD ANY OPTIONS TO IT
\title{Leaking Sensitive Financial Accounting Data \\ in Plain Sight using Deep Autoencoder Neural Networks}
\author{
    Marco Schreyer\textsuperscript{\rm 1}, 
    Christian Schulze\textsuperscript{\rm 2}, 
    Damian Borth\textsuperscript{\rm 1}\\
}
\affiliations{
    \textsuperscript{\rm 1}University of St. Gallen (HSG), St. Gallen, Switzerland\\
    \textsuperscript{\rm 2}German Research Center for Artificial Intelligence (DFKI), Kaiserslautern, Germany\\

    \{marco.schreyer, damian.borth\}@unisg.ch, christian.schulze@dfki.de

}
\begin{document}

\maketitle

\begin{abstract}
Nowadays, organizations collect vast quantities of sensitive information in `Enterprise Resource Planning' (ERP) systems, such as accounting relevant transactions, customer master data, or strategic sales price information. The leakage of such information poses a severe threat for companies as the number of incidents and the reputational damage to those experiencing them continue to increase. At the same time, discoveries in deep learning research revealed that machine learning models could be maliciously misused to create new attack vectors. Understanding the nature of such attacks becomes increasingly important for the (internal) audit and fraud examination practice. The creation of such an awareness holds in particular for the fraudulent data leakage using deep learning-based steganographic techniques that might remain undetected by state-of-the-art `Computer Assisted Audit Techniques' (CAATs). In this work, we introduce a real-world `threat model' designed to leak sensitive accounting data. In addition, we show that a deep steganographic process, constituted by three neural networks, can be trained to hide such data in unobtrusive `day-to-day' images. Finally, we provide qualitative and quantitative evaluations on two publicly available real-world payment datasets.
\end{abstract}

\section{Introduction}
\label{sec:introduction}

\noindent In recent years data became one of the organizations most valuable asset (`the new oil' \citep{economist2017world}). As a result, data protection, preventing it from being stolen or leaked to the outside world, is often considered of paramount importance. This observation holds in particular for data processed and stored in \textit{Enterprise Resource Planning} (ERP) systems. Steadily, these systems collect vast quantities of business process and accounting data at a granular level. Often the data recorded by such system encompasses sensitive information such as (i) the master data of customers and vendors, (ii) the volumes of sales and revenue, and (iii) strategic information about purchase and sales prices. Nowadays, the leakage of such information poses a severe issue for companies as the number of incidents and the costs to those experiencing them continue to increase\footnote{An overview of leakage incidents: \url{https://selfkey.org/data-breaches-in-2019/}.}. The potential damage and adverse consequences of a leakage incident may result in indirect losses, such as violations of (privacy) regulations or settlement and compensation fees. Furthermore, it can also result in indirect losses, such as damaging a company's goodwill and reputation or the exposure of the intellectual property. Formally, \textit{data leakage} can be defined as the \textit{`unauthorized transmission of information from inside an organization to an external recipient'} \citep{shabtai2012}. The key characteristic of a data leak incident is that it is carried out by `insiders' of an organization, e.g., employees or subcontractors. The detection of such insider attacks remains difficult since it often involves the misuse of legitimate credentials or authorizations to perform such an attack. 

Recent breakthroughs in artificial intelligence, and in particular deep neural networks \cite{lecun2015}, created advances across a diverse range of application domains such as image classification \cite{krizhevsky2012}, speech recognition \cite{saon2015}, language translation \cite{sutskever2014} and game-play \cite{silver2017}. Due to their broad application, these developments also raised awareness of the unsafe aspects of deep learning \cite{szegedy2013, goodfellow2014, papernot2017}, which become increasingly important for the (internal-) audit practice. The research of the potential impact of deep learning techniques in accounting and audit is still in an early stage \cite{sun2019, schreyer2017}. However, we believe that it is of vital relevance to understand how such techniques can be maliciously misused in this sphere to create new attack vectors \cite{ballet2019, schreyer2019}. This holds in particular for the fraudulent\footnote{According to \citep{garner2014} the term \textit{fraud} refers to the use of one's occupation for personal enrichment through the deliberate misuse or misapplication of the using organization's resources or assets, e.g. the booking of fictitious sales, fraudulent invoices, wrong recording of expenses.} data leakage using deep learning-based steganographic techniques that might remain undetected by state-of-the-art \textit{Computer Assisted Audit Techniques} (CAATs). 

\begin{figure*}[t!]
    \begin{center}
        \includegraphics[width=0.7\textwidth]{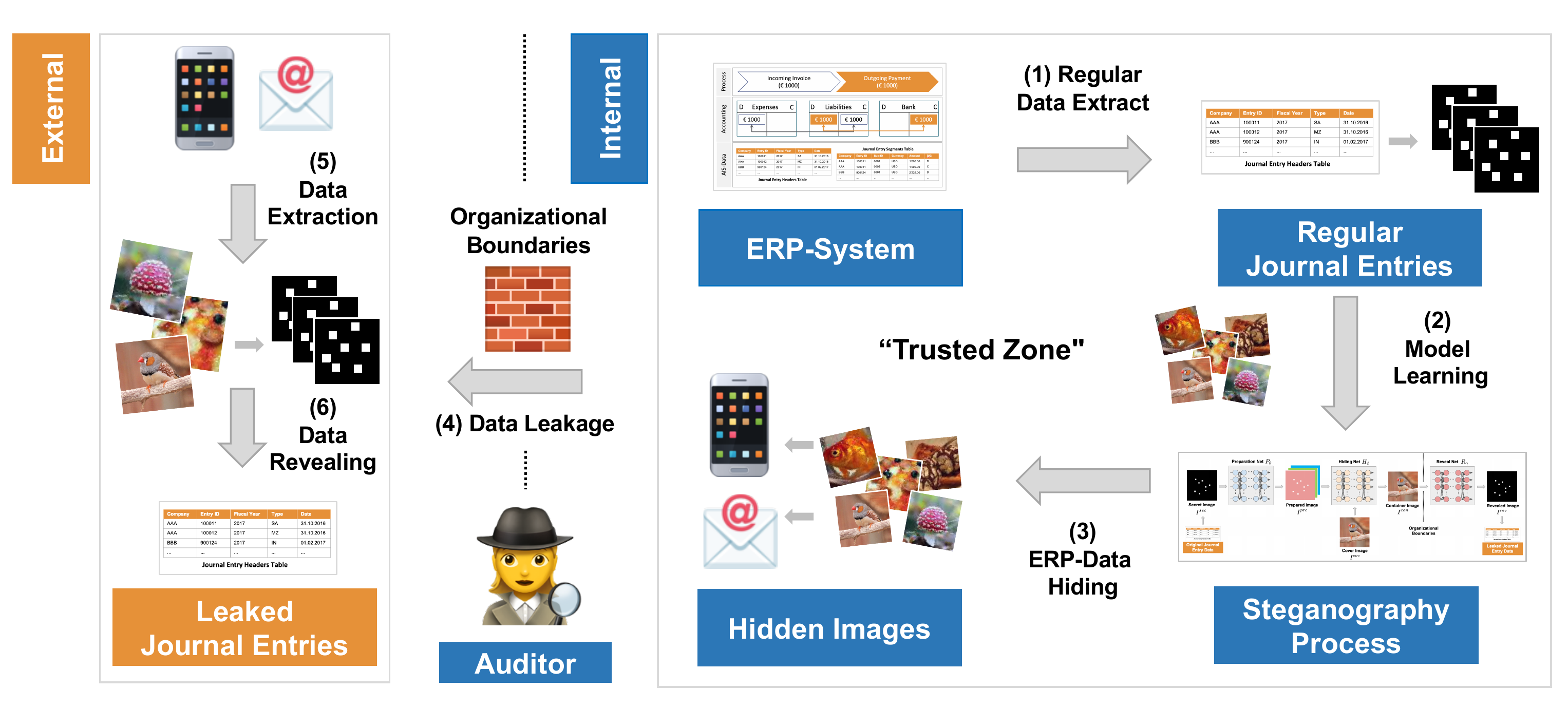}
    \end{center}
    \caption{Exemplary data leakage \textit{threat model} designed to leak sensitive accounting data such as (i) detailed journal entries or (ii) sensitive customer and vendor master data. The sensitive data is encoded into non-suspicious appearing `container' images that transmitted to a location outside of the organization.}
    \label{fig:threatmodel}
\end{figure*}

In general, the objective of \textit{steganography} denotes a secret communication \cite{shabtai2012}: a sender encodes a message into a cover medium, such as an image or audio file, such that the recipient can decode the message. At the same time, a potential (internal-) auditor or fraud examiner cannot judge whether the cover medium contains a secret message or not. In this paper, we present a deep learning-based steganographic process designed to leak sensitive accounting data. We regard this work to be an initial step towards the investigation of such future challenges of audits or fraud examinations. In summary, we present the following contributions: First, we describe a real-world `threat model' designed to leak sensitive accounting data using deep steganographic techniques. Second, we show that deep neural networks are capable of learning to hide the accounting data in everyday data, such as simple `cover' images. The created cover images are deliberately designed not to raise awareness and misguide auditors or fraud examiners. Third, to complete the attack scenario, we demonstrate how the hidden information can be revealed from the cover images upon the successful leak. % The remainder of this work is structured as follows: In Section \ref{sec:related_work}, we provide an overview of related work. Section \ref{sec:threat_model} follows with a description of the envisioned threat model and Section \ref{sec:methodology} presents the proposed methodology deep steganographic architecture. The experimental setup and results are outlined in Section \ref{sec:experiments} and Section \ref{sec:results}. In Section \ref{sec:conclusion}, the paper concludes with a summary of the current work and future research directions.

\section{Related Work}
\label{sec:related_work}

A wide variety of steganography settings and methods have been propsed in the literature; most relevant to our work are methods for blind image steganography, where the message is encoded in an image and the decoder does not have access to the original cover image. 

\textbf{Least Significant Bit (LSB) Methods:} The Least Significant Bit \cite{chandramouli2001, fridrich2001, mielikainen2006, viji2011, qazanfari2017} describes a classic steganographic algorithm. In general, each pixel of a digital image is comprised of three bytes (i.e., 8 binary bits) that represent the RGB chromatic values respectively. The \textit{n}bit-LSB algorithm replaces the least \textit{n} significant bits of the cover image by the \textit{n} most significant bits of the secret image. For each byte, the significant bits dominate the colour values. This way, the chromatic variation of the container image (altered cover) is minimized. Revealing the concealed secret image can be accomplished by reading the \textit{n} least significant bits and performing a bit shift despite that its distortion is often not visually observable. However, the LSB algorithm is, highly vulnerable to steganalysis. Often statistical analyses can easily detect the pattern of the altered pixels as shown in \cite{fridrich2001, lerch2016, luo2010}. Recent works have therefore focused on methods that preserve the original image statistics or the design of sophisticated distortion functions \cite{holub2012, holub2014, long2018, pevny2010, swain2018, tamimi2013}. 

\textbf{Discrete Cosine Transform (DCT) Methods:} To overcome the drawbacks of the LSB algorithm, the variant HRVSS \cite{eltahir2009} and \cite{muhammad2018} focus for example, on the unique biological trait of human eyes when hiding grey images in colour images. Other approaches utilize the bit-plane complexity segmentation in either the spatial or the transform domain \cite{kawaguchi1999, spaulding2002, ramani2007}. Other algorithms embed secret information in the Discrete Cosine Transformation `DCT' frequency domain of an image. This is achieved by deliberately changing the DCT coefficients \cite{chae1999, kaur2011, nag2011, zhang2018}. As many of those coefficients are zero in general, they can be exploited to embed information in the images frequency domain rather than the spatial domain \cite{elrahman2018, cheddad2010, sadek2015}. 

\textbf{Deep Learning Methods:} Recently, deep learning-based steganography methods have been proposed to encode (binary) text messages in images \cite{baluja2017, zhu2018}. Prior works \cite{husien2015, pibre2015} mostly focused on the decoding of the hidden information (in terms of determining which bits to extract from the container image) to increase the transmission accuracy. Besides, deep learning methods are also increasingly employed in the context of steganalysis to detect the information potentially hidden \cite{pibre2016, qian2015, tan2014, xu2016}. In this work, we build on the ideas of  \cite{baluja2017, hayes2017} that proposed an entire steganographic based on deep networks, encompassing a preparation, an encoding (hiding) and decoding (revealing) network. To the best of our knowledge, this work presents the first analysis of a deep neural network based data leakage attack targeting sensitive financial accounting data.

\section{Deep Steganography Threat Model}
\label{sec:threat_model}

To establish such a data leakage attack, we introduce a \textit{threat model} depicted in Fig. \ref{fig:threatmodel}, in which a person or group of people, referred to as the \textit{perpetrator}, within an organization intends to transmit `leak' sensitive data of the organization to an external destination or recipient. To initiate the attack, a perpetrator will query the ERP system to extract the sensitive accounting data to be leaked, such as detailed journal entry data or customer and vendor master data tables (1). Afterwards, the extracted entries are converted into binary images, referred to as \textit{secret images}, exhibiting a `one-hot' encoded representation of the data. One could think of converting the information into a Quick Response (QR) code representation using designated publicly available software\footnote{GitHub: \url{https://github.com/Bacon/BaconQrCode}}. 

Following, a population of images referred to as \textit{cover images} is selected. Usually, the cover images contain unobtrusive content to leak the secret information covertly. Such images can, for example, be obtained from prominent and publicly available image databases such us \cite{thomee2016, ILSVRC15}. Following, a deep \textit{steganographic model} is learned utilizing a deliberately designed training process. The process is comprised of a sequence of deep neural networks that process both the secret and the cover images. The model is trained to create a \textit{container image} that efficiently encodes the information contained in the secret image into the cover image (2). Thereby the encoding is regularized to hide the secret information with minimal distortion of the visual appearance of the cover image. Simultaneously, the model is trained to also decode the encoded secret information from the container image. 

Upon successful training, the model is exploited to facilitate the attack (3). Therefore, a single or a series of container images are created that conceal the sensitive data to be leaked. The created container images are then transmitted either (i) electronically (e.g., via email or files-haring) or (ii) physically (e.g., via mobile phone or USB flash drive) to a location outside of the organizational boundaries, usually referred to as `safe zone'. (4). Visually the container images seem to correspond to unobtrusive `everyday' images but encode the sensitive information to be leaked. As a result, they exhibit a high chance of being judged as non-suspicious throughout an internal audit. Once the container images are successfully transmitted, the secret binary images can be revealed using the steganographic model (5). Ultimately the sensitive data can be encoded at high accuracy (6). 

\section{Deep Steganography Architecture}
\label{sec:methodology}

\begin{figure*}[t!]
    \begin{center}
        \includegraphics[width=0.9\textwidth]{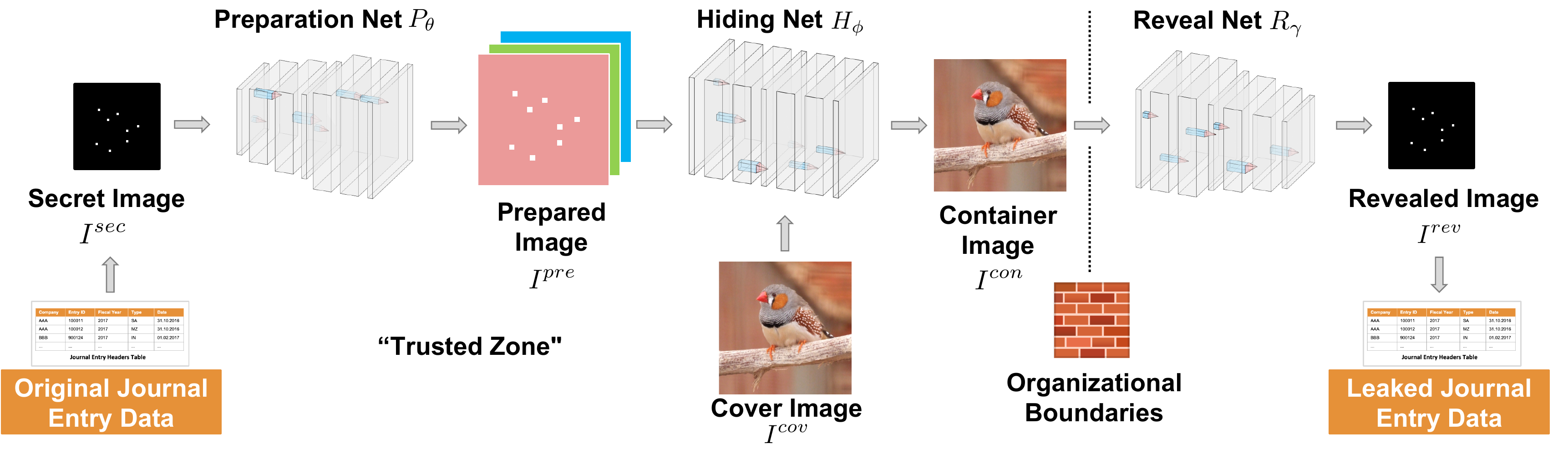}
    \end{center}
    \caption{The deep architecture as introduced in \citep{baluja2017}, applied to learn a steganographic model of real-world accounting data. The architecture is designed to encode and decode sensitive information into unobtrusive `day-to-day' cover images.}
    \label{fig:architecture}
\end{figure*}

Inspired by and building on the work of Baluja \cite{baluja2017} we investigate if a deep-learning based steganography process can be misused to leak sensitive accounting data. The proposed architecture, is constituted of three neural networks: a \textit{preparation} network $P_\theta$, a \textit{hiding} network $H_\phi$, and a \textit{reveal} network $R_\gamma$. The architecture is trained `end-to-end' and $\theta$, $\phi$, and $\gamma$ denote the trainable parameters of each network respectively. 

To establish a potential data leak, a set of $X$ of $N$ journal entries $x^{1}, x^{2}, ..., x^{n}$ (or other tabular data record) is extracted from the ERP system to be attacked. Each journal entry $x^{i}$ consists of $M$ accounting specific attributes $x_{1}^{i}, x_{2}^{i}, ..., x_{j}^{i}, ..., x_{m}^{i}$. The individual attributes $x_{j}$ describe the journal entries details, e.g., the entries' fiscal year, posting type, posting date, amount, general-ledger. Afterwards, each entry is sequentially processed by the distinct networks as described in the following. 

\textbf{Prepare Step:}  Each entry (or multiple entries) is converted into a \textit{secret image}. The secret image $I^{sec} \in \{0,1\}^{H \times W}$ of shape $H \times W$ contains a binary representation of the information to be leaked. The preparation network $P_{\theta}$ receives $I^{sec}$ and converts it into a RGB \textit{prepared image} $I^{pre} \in \{0,255\}^{C \times H \times W}$ that exhibits the desired characteristics to be hidden. 

\textbf{Hide Step:} The hiding process is then initiated using the prepared image and a cover image. The RGB cover image $I^{cov} \in \{0,255\}^{C \times H \times W}$ of shape $C \times H \times W$ corresponds to an arbitrary `everyday' image and is used to hide the prepared image. The hiding network $H_\phi$ receives a cover image $I^{cov}$ and the prepared image $I^{pre}$. Using both input images, $H_\phi$ produces a RGB \textit{container image} $I_{con}$ of the same shape as $I^{cov}$. 

\textbf{Reveal Step:} Subsequently, the container image is passed to the reveal network $R_\gamma$ to produce a binary reveal image $I^{rev} \in \{0,1\}^{H \times W}$ that exhibits a high similarity to the original prepared image $I^{pre}$. Ultimately, $I^{rev}$ is supposed to reveal the secret information of $I^{sec}$ hidden in $I^{con}$. The training of the architecture follows a dual training objective, namely (1) the minimization of the cover distortion (\textit{cover loss}) and (2) the preservation of the secret information (\textit{secret loss}) when optimizing the network parameters. 

% %Thereby, $H_\phi$ encodes $I^{pre}$ into $I^{cov}$ with a minimal distortion of the visual appearance of the cover image.

% Todo: extract can be derived from database or screen 

\textbf{Cover loss:} The first training objective requires that the container image $I^{con}$ exhibits a high visual similarity to the cover image $I^{cov}$. Formally defined as the mean-squared difference $\mathcal{L}^{\scaleto{MSE}{3pt}}_{\theta, \phi}$ between both images, given by:

\begin{equation}
	\mathcal{L}^{\scaleto{MSE}{3pt}}_{\theta, \phi} = \sum_{x=0}^{C-1} \sum_{y=0}^{H-1} \sum_{z=0}^{W-1} (I^{cov}_{x, y, z} - I^{con}_{x, y, z})^{2}, 
\label{equ:mse_loss}
\end{equation}

\noindent where $C$, $H$, and $W$ denote the shape of both RGB images and $\theta$ and $\phi$ the network parameters. 

\textbf{Secret loss:} The second training objective requires that the revealed image $I^{rev}$ should contain the same binary information as encoded in the original secret image $I^{sec}$. Formally defined as the binary cross-entropy difference $\mathcal{L}^{\scaleto{BCE}{3pt}}_{\theta, \phi, \gamma}$ of between images, given by:

\begin{equation}
\begin{aligned}
	\mathcal{L}^{\scaleto{BCE}{3pt}}_{\theta, \phi, \gamma} = {} 
	& \sum_{x=0}^{H-1} \sum_{y=0}^{W-1} I^{sec}_{x, y} \log I^{rev}_{x, y} \\ 
	& + (1 - I^{sec}_{x, y}) \log (1 - I^{rev}_{x, y}), 
\end{aligned}
\label{equ:bce_loss}
\end{equation}

\noindent where $H$, and $W$ denote the shape of both binary images and $\theta$, $\phi$ and $\gamma$ the network parameters. We train the network parameters $\theta$, $\phi$, and $\gamma$ batch-wise using stochastic gradient descent to minimize a combined loss function over the distributions of secret and cover images, as defined by: 

\begin{equation}
	\mathcal{L}^{\scaleto{ALL}{3pt}}_{\theta, \phi, \gamma} = \alpha \cdot \frac{1}{N} \sum_{i=0}^{N-1} \mathcal{L}^{\scaleto{MSE}{3pt}}_{\theta, \phi} + \beta \cdot \frac{1}{N} \sum_{i=0}^{N-1} \mathcal{L}^{\scaleto{BCE}{3pt}}_{\theta, \phi, \gamma}, 
\label{equ:combined_loss}
\end{equation}

\noindent where $N$ denotes the size of the training batch and the factors $\alpha$ and $\beta$ balance both losses. In summary, the approach builds upon the idea of autoencoder neural networks originally proposed in \cite{hinton2006}. We encoded two input images such that the learned intermediate representation $I^{con}$ comprises an image that appears as similar as possible $I^{cov}$ and thereby hides the accounting data information contained in $I^{sec}$.

\section{Datasets and Experimental Setup}
\label{sec:experiments}

% \noindent \textbf{Datasets and Data Preprocessing:} 

Due to the high confidentiality of real-world accounting data and to allow for reproducibility of our results, we evaluate the proposed methodology based on two publicly available datasets of real-world city payments. The payment datasets serve as \textit{secret data} to be hidden. We use a prominent dataset of everyday images that serves as \textit{cover data} to hide the payment information.   

\textbf{Secret Data:} The secret datasets to be leaked, encompass two publicly available payment datasets of the city of Philadelphia and the city of Chicago. The majority of attributes recorded in both datasets (similar to real-world ERP data) correspond to categorical (discrete) variables, e.g. posting date, department, vendor name, document type. For each dataset we pre-process the original payment line-item attributes to (i) remove semantically redundant attributes and (ii) obtain a binary (`one-hot') representation of each payment record. The following descriptive statistics summarise both datasets upon successful data pre-processing:

The \textbf{`City of Philadelphia'} dataset\footnote{ https://www.phila.gov} (\textit{dataset A}) encompasses the city's payments of the fiscal year 2017. It represents nearly \$4.2 billion in payments obtained from almost 60 city offices, departments, boards and committees. The dataset encompasses $n=238,894$ payments comprised of $10$ categorical and one numerical attribute. The encoding resulted in a total of $\mathcal{D} = 8,565$ one-hot encoded dimensions for each vendor payment record $x^{i} \in \mathcal{R}^{8,565}$.

The \textbf{`City of Chicago'} dataset\footnote{ https://data.cityofchicago.org} (\textit{dataset B}) encompasses the city's vendor payments ranging from 1996 to 2020. The data is collected from the city's `Vendor, Contract and Payment Search' and encompasses the procurement of goods and services. The dataset encompasses $n=72,814$ payments comprised of $7$ categorical and one numerical attribute. The encoding resulted in a total of $\mathcal{D} = 2,354$ one-hot encoded dimensions for eachvendor payment record $x^{i} \in \mathcal{R}^{2,354}$.

\textbf{Cover Data:} The cover dataset used to hide the secret data is derived from the `ImageNet 2012' dataset. The `ImageNet 2012' \cite{ILSVRC15} dataset\footnote{http://image-net.org/download-images} (\textit{ImageNet}) is a publicly available datasets commonly used to evaluate machine learning models. The original dataset contains a total of 1.2 million RGB images organized in non-overlapping 1,000 subcategories. Each category corresponds to a real-world concept, such as `llama', `guitar', or `pillow'. We use a subset of the entire dataset referred to as the `Tiny ImageNet' dataset\footnote{https://tiny-imagenet.herokuapp.com}. The dataset consists of 200 ImageNet subcategories. Thereby, each category encompasses 500 images resulting in a total of 100k RGB images. To demonstrate the generalization of the proposed method, we crop from each image a random image patch of size 224 $\times$ 224 pixels per channel and resize it to 256 $\times$ 256 pixels.

\begin{table*}[!ht]
 \caption{Network training losses ($\mathcal{L}^{\scaleto{ALL}{3pt}}_{\theta, \phi, \gamma}$, $\mathcal{L}^{\scaleto{MSE}{3pt}}_{\theta, \phi}$, $\mathcal{L}^{\scaleto{BCE}{3pt}}_{\theta, \phi, \gamma}$), PSNR, SSIM, and BACC obtained on both city payment datasets when using three channel RGB ($H\!\times\!W\!\times\!3$) cover images.} 
 \centering
 \fontsize{8}{7}\selectfont
 \input{results_a.tex}
 \label{tab:quantitative_results_a}
\end{table*}

\begin{table*}[!ht]
 \caption{Network training losses ($\mathcal{L}^{\scaleto{ALL}{3pt}}_{\theta, \phi, \gamma}$, $\mathcal{L}^{\scaleto{MSE}{3pt}}_{\theta, \phi}$, $\mathcal{L}^{\scaleto{BCE}{3pt}}_{\theta, \phi, \gamma}$), PSNR, SSIM, and BACC obtained on both city payment datasets when using single channel grayscale ($H\!\times\!W\!\times\!1$) cover images.} 
 \centering
 \fontsize{8}{7}\selectfont
 \input{results_b.tex}
 \label{tab:quantitative_results_b}
\end{table*}

%\subsection{Experimental Setup}
%\label{sec:setup}

\textbf{Architectural Setup:} The steganographic process as shown in Fig. \ref{fig:architecture} and described in \cite{baluja2017} is trained end-to-end comprised of the \textit{preparation} network $P_\theta$, the \textit{hiding} network $H_\phi$, and the \textit{reveal} network $R_\gamma$. Each network applies a series 2D convolutions \cite{lecun1995} of different filter sizes applied onto the input images (architectural details are described in the appendix). The convolutions are followed by non-linear $ReLU$ activation's \cite{nair2010} except for the three output layers of the preparation network were $Sigmoid$ activation's are used. 

% $Conv^{c}_{k \times k}$ \cite{lecun1995} of different filter sizes applied onto the input images, where $c$ denotes the number of output channels and $k \times k$ denotes the filter kernel size.

\textbf{Training Setup:} The networks are trained on secret images $I^{sec}$ derived from the entire population of payments in each dataset. Throughout the training, the secret images are paired with randomly drawn cover images $I^{cov}$ of the ImageNet dataset. We train with a mini-batch size of $m=6$ secret-cover image pairs for a max. $\tau=$ 800k iterations and apply early stopping once the combined loss defined in Eq. \ref{equ:combined_loss} converges. Thereby, the network parameters $\theta$, $\phi$, and $\gamma$ are optimized with a constant learning rate of $\eta=10^{-5}$ using Adam optimization \cite{kingma2014}, setting $\beta_{1}=0.9$ and $\beta_{2}=0.999$. We sweep the weight factor of the container image loss $\mathcal{L}^{\scaleto{MSE}{3pt}}_{\theta, \phi}$ through $\alpha \in [0.2, 1.0]$ to determine an optimal hyperparameter setup.

\section{Experimental Results}
\label{sec:results}

\begin{figure*}[!ht]
 \centering
 \input{results_c.tex}
 \caption{Exemplary data hiding results of a deep steganographic architecture model trained for $\tau\!=\!800k$ iterations using RGB cover images (left to right): original secret, revealed secret, original cover, and created container image, as well as the pixel-wise residual errors of the cover and the container image.}
 \label{fig:qualitative_results_a}
\end{figure*}
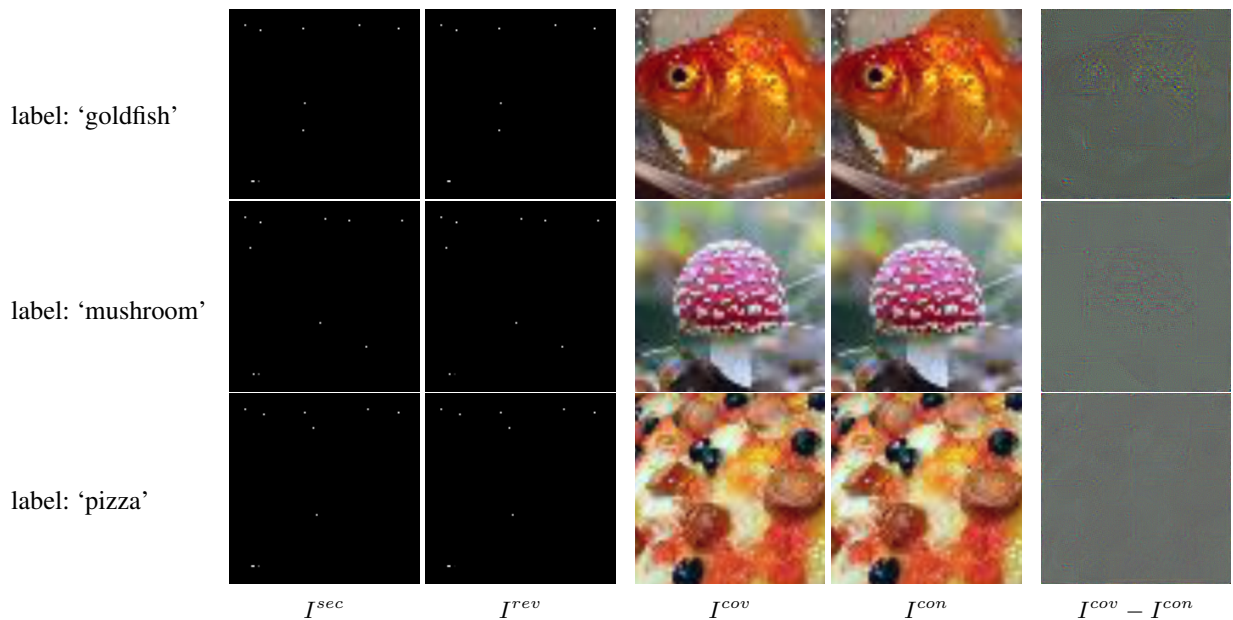

\begin{figure*}[!ht]
 \centering
 \input{results_d.tex}
 \caption{Exemplary data hiding results using grayscale cover images (left to right): original secret, revealed secret, prepared secret, original cover, and created container image, as well as the pixel-wise residual errors of the cover and the container image.}
 \label{fig:qualitative_results_b}
\end{figure*}
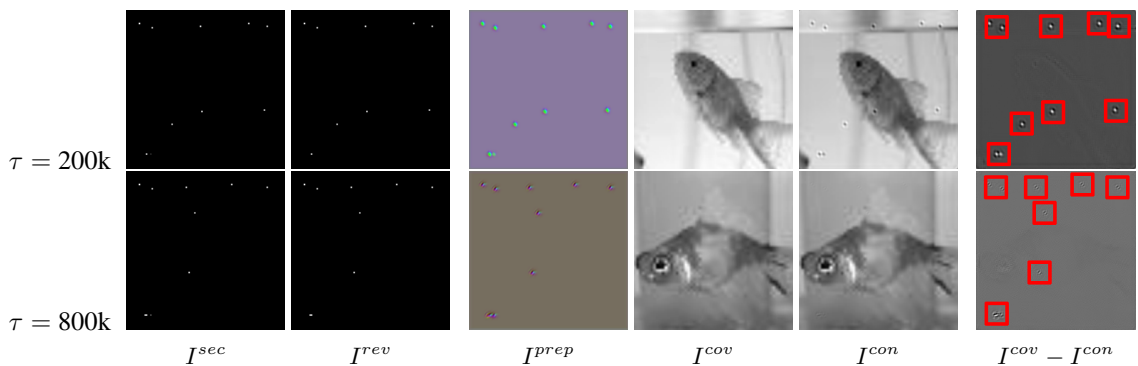

In this section, we quantitatively and qualitatively assess the data leakage capability of the introduced steganographic process. The quantitative evaluation is conducted according to: (i) model \textit{secrecy}, the difficulty of detecting the sensitive accounting data hidden in each image; (ii) model \textit{accuracy}, the extent to which the secret information can be revealed correctly from the container image. % upon a successful leak. 

\textbf{Quantitative Evaluation:} We evaluate the model secrecy using the \textit{Peak Signal-to-Noise Ratio} (PSNR) ratio as a proxy. The PSNR measures the ratio between the maximum possible power of a signal and the power of corrupting noise that affects the fidelity of its representation, as defined by: 

\begin{equation}
    \hbox{PSNR} = 10 \cdot \log_{10} \frac{\max | I^{cov} |}{\mathcal{L}^{\scaleto{MSE}{3pt}}_{\theta, \phi}(I^{cov}, I^{con})},
\label{equ:psnr}
\end{equation}

\noindent where $\max |I^{cov}|$ denotes the maximum absolute possible value of the cover image. In the absence of noise, the cover image $I^{cov}$ and the container image $I^{con}$ are identical, and thus the $\mathcal{L}^{\scaleto{MSE}{3pt}}_{\theta, \phi}$ is zero. In this case, the PSNR is infinite \cite{salomon2004}. It is assumed that acceptable values for wireless transmission quality loss are considered to be about 20 dB to 25 dB \cite{li2007}. In addition, we measure the \textit{Structural Similarity Index Measure (SSIM)} \cite{wang2004}. The SSIM defines a perception-based model considering image degradation as a perceived change in structural information. We quantify the visibility of differences between the distorted container image $I^{con}$ and the cover image $I^{cov}$, as defined by:

\begin{equation}
    \hbox{SSIM} = \frac {(2\mu _{cov}\mu _{con}+c_{1})(2\sigma _{cov, con}+c_{2})}{(\mu _{cov}^{2}+\mu _{con}^{2}+c_{1})(\sigma _{cov}^{2}+\sigma _{con}^{2}+c_{2})},
\label{equ:ssim}
\end{equation}

\noindent where $\mu _{cov}$ and $\mu _{con}$ denotes the average pixel value of $I^{cov}$ and $I^{con}$ respectively, $\sigma _{cov}$ and $\sigma _{con}$ their variance, and $\sigma _{cov, con}$ the co-variance of the pixel values in both images. The SSIM maximum possible value of 1 indicates that the two images are structurally similar, while a value of 0 indicates no structural similarity. 

\noindent We measure the model accuracy using \textit{Bit Accuracy (ACC)}, which denotes the fraction of identical active bits between the secret image and revealed image, as defined by: 

\begin{equation}
    \hbox{BACC} = 1 - \frac{\sum_{x=0}^{H-1} \sum_{y=0}^{W-1} \mathbbm{1}_{[\|I^{sec}_{x, y} - M \circ I^{rev}_{x, y}\| \leq \delta]}}{\sum_{x=0}^{H-1} \sum_{y=0}^{W-1} \mathbbm{1}_{[I^{sec}_{x, y} > 0]}},
\label{equ:bacc}
\end{equation}

\noindent where $\mathbbm{1}$ denotes the indicator function and $M= \max (I^{rev}, k)$ denotes a binary pixel mask corresponding to the $k$ most active pixel values in $I^{rev}$. We set $k$ to the number of active binary pixels in $I^{sec}$. The operator $\circ$ denotes the element-wise multiplication, and $\delta$ denotes the dissimilarity threshold of the most active pixel values. We set to $\delta=0.001$ in all our experiments.

We analyze the performance of the introduced process in two setups: In the first setup, we use the original three-channel RGB images of the ImageNet dataset to hide the secret information. In the second setup, we reduced the capacity of the cover images by converting the dataset of cover images to single-channel grayscale images. We measure capacity as \textit{Bits Per Pixel (BPP)} \cite{zhu2018}, which denotes the number of encoded secret bits in the encoded image, defined by $\mathcal{D} / (H\!\times\!W\!\times\!C)$. The trained models are evaluated on randomly drawn combinations of 5k secret-cover image pairs. Tables \ref{tab:quantitative_results_a} and \ref{tab:quantitative_results_b} show the quantitative results of both setups. It can be observed that in the RGB cover image setup different $\alpha$ parameterizations yield high a secrecy and accuracy of the secret information. Similar results are obtained for the grayscale cover image setup, indicating that the grayscale images provide sufficient capacity to hide the data to the unaided human eye. 

\textbf{Qualitative Evaluation:} Figure \ref{fig:qualitative_results_a} shows three exemplary data hiding results using RGB cover images. The reconstructed container images $I^{con}$ appear almost identical to the original cover images $I^{cov}$ and encode all the information necessary to reconstruct the original payment information. The rightmost column shows the pixel-wise residual error between $I^{cov}$ and $I^{con}$. It can be observed that the secret information is encoded across the pixel intensities of $I^{con}$ and cannot easily be obtained. This learned obfuscation is of high relevance in a scenario when the original cover image becomes accessible to the auditor or fraud examiner. Figure \ref{fig:qualitative_results_b} shows two exemplary data hiding results of the grayscale cover image setup. It can be observed that, due to the limited capacity of the cover, the secret information remains partially visible in the pixel-wise residual error even when training the model for $\tau=800k$ iterations.
\label{subsec:qualitative}

\section{Conclusion}
\label{sec:conclusion}

In this work, we conducted an analysis of deep steganography as a future challenge to the (internal) audit and fraud examination practice. We introduced a `threat model' to leak sensitive information recorded in ERP systems. We also provided initial evidence that deep steganographic techniques can be maliciously misused to hide such information in unobtrusive `everyday' images. In summary, we believe that such an attack vector pose a substantial challenge for CAATs used by auditors nowadays and the near future.

\bibliography{library}

\appendix
\onecolumn
\raggedright
\textbf{Appendix A: Architectural Details}

\vspace*{3mm}

\begin{table}[!ht]
\caption{Architectural details of the distinct neural networks that constitute the steganography architecture, namely preparation network $P_\theta$, hiding network $H_{\phi}$, and reveal network $R_{\gamma}$.}
 \centering
 \input{architecture_details.tex}
 \label{tab:architecture}
\end{table}

\begin{justify} 
Each network applies a series of convolutions and non-linear activation's onto their respective input images. Each network applies a series of 2D convolutions $Conv^{c}_{k \times k}$ \cite{lecun1995} of different filter sizes applied onto the input images, where $c$ denotes the number of output channels and $k \times k$ denotes the filter kernel size. The convolutions are followed by non-linear $ReLU$ activation's \cite{nair2010} except for the three output layers of the preparation network were $Sigmoid$ activation's are used.
\end{justify}

\end{document}

%% file: results_a.tex
\begin{tabular}{c c c c | r r r | r r r  }

\multicolumn{1}{l}{}
& \multicolumn{1}{c}{}
& \multicolumn{1}{c}{}
& \multicolumn{1}{c}{}
\\
%\addlinespace[0.05cm]
\\
\multicolumn{1}{l}{Dataset}
& \multicolumn{1}{c}{$\hbox{BPP}$}
& \multicolumn{1}{c}{$\alpha$}
& \multicolumn{1}{c}{$\beta$}
& \multicolumn{1}{c}{$\mathcal{L}^{\scaleto{ALL}{3pt}}_{\theta, \phi, \gamma}$}
& \multicolumn{1}{c}{$\mathcal{L}^{\scaleto{MSE}{3pt}}_{\theta, \phi}$}
& \multicolumn{1}{c}{$\mathcal{L}^{\scaleto{BCE}{3pt}}_{\theta, \phi, \gamma}$}
& \multicolumn{1}{c}{$\hbox{PSNR}$}
& \multicolumn{1}{c}{$\hbox{SSIM}$}
& \multicolumn{1}{c}{$\hbox{BACC}$}
\\
\midrule
 \multirow{4}{*}{A} & \multirow{4}{*}{0.0436} & 0.2 & 1.0 & 0.65 $\pm$ 0.02 & 0.58 $\pm$ 0.12 & 0.54 $\pm$ 0.01 & 42.71 $\pm$ 0.98 & 0.996 $\pm$ 0.001 & 0.999 $\pm$ 0.001 \\
 & & 0.5 & 1.0 & 0.77 $\pm$ 0.07 & 0.57 $\pm$ 0.07 & 0.54 $\pm$ 0.04 & 42.95 $\pm$ 0.94 & 0.997 $\pm$ 0.001 & 0.998 $\pm$ 0.002 \\
 & & 0.8 & 1.0 & 0.93 $\pm$ 0.01 & 0.48 $\pm$ 0.05 & 0.55 $\pm$ 0.01 & 43.43 $\pm$ 0.45 & 0.996 $\pm$ 0.001 & 0.997 $\pm$ 0.001 \\
 & & 1.0 & 1.0 & 0.97 $\pm$ 0.06 & 0.53 $\pm$ 0.07 & 0.54 $\pm$ 0.02 & 41.87 $\pm$ 0.27 & 0.996 $\pm$ 0.001 & 0.998 $\pm$ 0.004 \\
 \midrule
 \multirow{4}{*}{B} & \multirow{4}{*}{0.0120} & 0.2 & 1.0 & 0.67 $\pm$ 0.07 & 0.79 $\pm$ 0.03 & 0.51 $\pm$ 0.00 & 41.38 $\pm$ 0.32 & 0.998 $\pm$ 0.002 & 0.999 $\pm$ 0.001 \\
 & & 0.5 & 1.0 & 0.77 $\pm$ 0.01 & 0.52 $\pm$ 0.02 & 0.51 $\pm$ 0.01 & 42.94 $\pm$ 0.23 & 0.996 $\pm$ 0.001 & 0.998 $\pm$ 0.001 \\
 & & 0.8 & 1.0 & 0.87 $\pm$ 0.05 & 0.46 $\pm$ 0.07 & 0.51 $\pm$ 0.00 & 43.51 $\pm$ 0.65 & 0.995 $\pm$ 0.003 & 0.998 $\pm$ 0.002 \\
 & & 1.0 & 1.0 & 0.99 $\pm$ 0.04 & 0.47 $\pm$ 0.04 & 0.52 $\pm$ 0.01 & 43.35 $\pm$ 0.37 & 0.997 $\pm$ 0.001 & 0.997 $\pm$ 0.003 \\
\midrule
\multicolumn{10}{l}{\scalebox{.9}{Values of the distinct loss functions $\mathcal{L}$ are multiplied by $10^{4}$, variances originate from parameter initialization using four distinct random seeds.}}
%\bottomrule \\
\end{tabular}

%% file: results_b.tex
\begin{tabular}{c c c c | r r r | r r r  }

\multicolumn{1}{l}{}
& \multicolumn{1}{c}{}
& \multicolumn{1}{c}{}
& \multicolumn{1}{c}{}
\\
%addlinespace[0.05cm]
\\
\multicolumn{1}{l}{Dataset}
& \multicolumn{1}{c}{$\hbox{BPP}$}
& \multicolumn{1}{c}{$\alpha$}
& \multicolumn{1}{c}{$\beta$}
& \multicolumn{1}{c}{$\mathcal{L}^{\scaleto{ALL}{3pt}}_{\theta, \phi, \gamma}$}
& \multicolumn{1}{c}{$\mathcal{L}^{\scaleto{MSE}{3pt}}_{\theta, \phi}$}
& \multicolumn{1}{c}{$\mathcal{L}^{\scaleto{BCE}{3pt}}_{\theta, \phi, \gamma}$}
& \multicolumn{1}{c}{$\hbox{PSNR}$}
& \multicolumn{1}{c}{$\hbox{SSIM}$}
& \multicolumn{1}{c}{$\hbox{BACC}$}
\\
\midrule
 A & 0.1307 & 0.5 & 1.0 & 0.55 $\pm$ 0.02 & 0.03 $\pm$ 0.02 & 0.54 $\pm$ 0.01 & 55.46 $\pm$ 0.67 & 0.998 $\pm$ 0.001 & 0.991 $\pm$ 0.001 \\
 B & 0.0359 & 0.5 & 1.0 & 0.87 $\pm$ 0.04 & 0.64 $\pm$ 0.07 & 0.55 $\pm$ 0.02 & 42.27 $\pm$ 0.49 & 0.989 $\pm$ 0.003 & 0.977 $\pm$ 0.002 \\
 \midrule
\multicolumn{10}{l}{\scalebox{.9}{Values of the distinct loss functions $\mathcal{L}$ are multiplied by $10^{4}$, variances originate from parameter initialization using four distinct random seeds.}}
%\bottomrule \\
\end{tabular}

%% file: results_c.tex
% row 1
\begin{minipage}[c][][c]{8em}
    \begin{flushleft}label: `goldfish' \end{flushleft} % (n01443537)
    \vspace{18mm}
\end{minipage}
\vspace{-10.6mm}
\begin{subfigure}[t]{0.30\columnwidth}
	\centering\includegraphics[width=\textwidth]{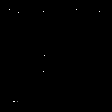}
\end{subfigure}
\begin{subfigure}[t]{0.30\columnwidth}
    \centering\includegraphics[width=\textwidth]{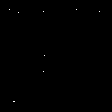}
\end{subfigure}
\begin{subfigure}[t]{0.01\columnwidth}
    \centering \hspace{1pt}
\end{subfigure}
\begin{subfigure}[t]{0.30\columnwidth}
    \centering\includegraphics[width=\textwidth]{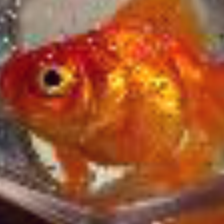}
\end{subfigure}
\begin{subfigure}[t]{0.30\columnwidth}
    \centering\includegraphics[width=\textwidth]{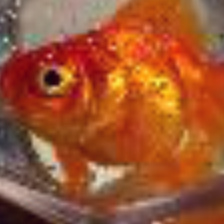}
\end{subfigure}
\begin{subfigure}[t]{0.01\columnwidth}
    \centering \hspace{1pt}
\end{subfigure}
\begin{subfigure}[t]{0.30\columnwidth}
    \centering\includegraphics[width=\textwidth]{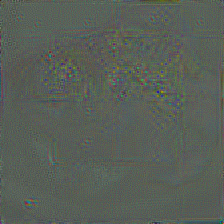}
\end{subfigure} \\
% row 2
\begin{minipage}[c][][c]{8em}
    \begin{flushleft}label: `mushroom' \end{flushleft} % (n07734744)
    \vspace{18mm}
\end{minipage}
\vspace{-10.2mm}
\begin{subfigure}[t]{0.30\columnwidth}
	\centering\includegraphics[width=\textwidth]{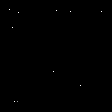}
\end{subfigure}
\begin{subfigure}[t]{0.30\columnwidth}
    \centering\includegraphics[width=\textwidth]{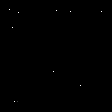}
\end{subfigure}
\begin{subfigure}[t]{0.01\columnwidth}
    \centering \hspace{1pt}
\end{subfigure}
\begin{subfigure}[t]{0.30\columnwidth}
    \centering\includegraphics[width=\textwidth]{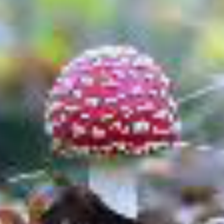}
\end{subfigure}
\begin{subfigure}[t]{0.30\columnwidth}
    \centering\includegraphics[width=\textwidth]{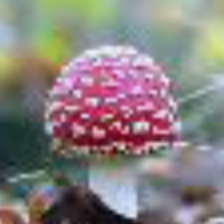}
\end{subfigure}
\begin{subfigure}[t]{0.01\columnwidth}
    \centering \hspace{1pt}
\end{subfigure}
\begin{subfigure}[t]{0.30\columnwidth}
    \centering\includegraphics[width=\textwidth]{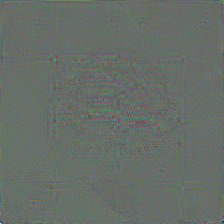}
\end{subfigure} \\
% row 3
\begin{minipage}[c][][c]{8em}
    \begin{flushleft}label: `pizza' \end{flushleft} % (n07873807)
    \vspace{18mm}
\end{minipage}
\vspace{-10.5mm}
\begin{subfigure}[t]{0.30\columnwidth}
	\centering\includegraphics[width=\textwidth]{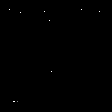}
	\caption*{$I^{sec}$}
\end{subfigure}
\begin{subfigure}[t]{0.30\columnwidth}
    \centering\includegraphics[width=\textwidth]{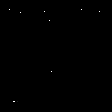}
    \caption*{$I^{rev}$}
\end{subfigure}
\begin{subfigure}[t]{0.01\columnwidth}
    \centering \hspace{1pt}
\end{subfigure}
\begin{subfigure}[t]{0.30\columnwidth}
    \centering\includegraphics[width=\textwidth]{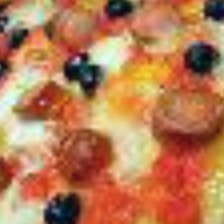}
    \caption*{$I^{cov}$}
\end{subfigure}
\begin{subfigure}[t]{0.30\columnwidth}
    \centering\includegraphics[width=\textwidth]{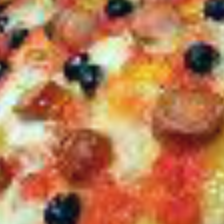}
    \caption*{$I^{con}$}
\end{subfigure}
\begin{subfigure}[t]{0.01\columnwidth}
    \centering \hspace{1pt}
\end{subfigure}
\begin{subfigure}[t]{0.30\columnwidth}
    \centering\includegraphics[width=\textwidth]{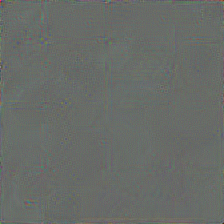}
    \caption*{$I^{cov} - I^{con}$}
\end{subfigure} \\
\vspace{6mm}

%% file: results_d.tex
% row 1
\centering
$\tau=$ 200k \hspace{1pt}
\begin{subfigure}[t]{0.25\columnwidth}
    \centering\includegraphics[width=\textwidth]{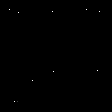}
\end{subfigure}
\begin{subfigure}[t]{0.25\columnwidth}
    \centering\includegraphics[width=\textwidth]{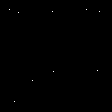}
\end{subfigure}
\begin{subfigure}[t]{0.01\columnwidth}
    \centering \hspace{1pt}
\end{subfigure}
\begin{subfigure}[t]{0.25\columnwidth}
    \centering\includegraphics[width=\textwidth]{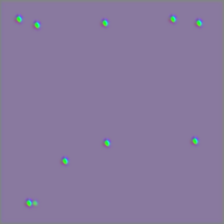}
\end{subfigure}
\begin{subfigure}[t]{0.25\columnwidth}
    \centering\includegraphics[width=\textwidth]{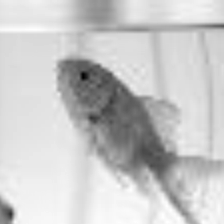}
\end{subfigure}
\begin{subfigure}[t]{0.25\columnwidth}
    \centering\includegraphics[width=\textwidth]{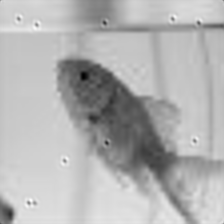}
\end{subfigure}
\begin{subfigure}[t]{0.01\columnwidth}
    \centering \hspace{1pt}
\end{subfigure}
\begin{subfigure}[t]{0.25\columnwidth}
    \centering\includegraphics[width=\textwidth]{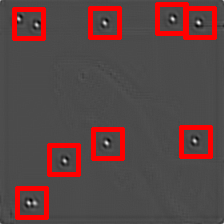}
\end{subfigure} \\
% row 2
\centering
$\tau=$ 800k \hspace{1pt}
\begin{subfigure}[t]{0.25\columnwidth}
	\centering\includegraphics[width=\textwidth]{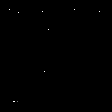}
	\caption*{$I^{sec}$}
\end{subfigure}
\begin{subfigure}[t]{0.25\columnwidth}
    \centering\includegraphics[width=\textwidth]{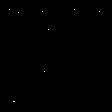}
    \caption*{$I^{rev}$}
\end{subfigure}
\begin{subfigure}[t]{0.01\columnwidth}
    \centering \hspace{1pt}
\end{subfigure}
\begin{subfigure}[t]{0.25\columnwidth}
    \centering\includegraphics[width=\textwidth]{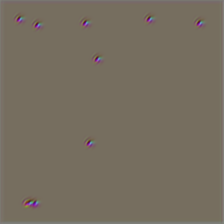}
    \caption*{$I^{prep}$}
\end{subfigure}
\begin{subfigure}[t]{0.25\columnwidth}
    \centering\includegraphics[width=\textwidth]{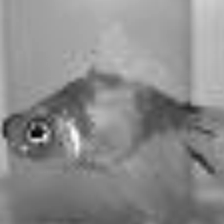}
    \caption*{$I^{cov}$}
\end{subfigure}
\begin{subfigure}[t]{0.25\columnwidth}
    \centering\includegraphics[width=\textwidth]{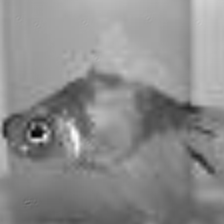}
    \caption*{$I^{con}$}
\end{subfigure}
\begin{subfigure}[t]{0.01\columnwidth}
    \centering \hspace{1pt}
\end{subfigure}
\begin{subfigure}[t]{0.25\columnwidth}
    \centering\includegraphics[width=\textwidth]{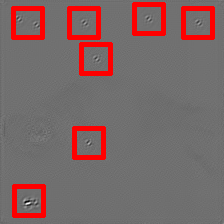}
    \caption*{$I^{cov} - I^{con}$}
\end{subfigure} \\
% row 3
%\centering
%$\tau=$ 150k \hspace{1pt}
%\begin{subfigure}[t]{0.13\columnwidth}
%	\centering\includegraphics[width=\textwidth]{01_secret_image_003.png}
%	\caption*{$I^{sec}$}
%\end{subfigure}
%\begin{subfigure}[t]{0.13\columnwidth}
%    \centering\includegraphics[width=\textwidth]{03_revealed_image_003.png}
%    \caption*{$I^{rev}$}
%\end{subfigure}
%\begin{subfigure}[t]{0.01\columnwidth}
%    \centering \hspace{1pt}
%\end{subfigure}
%\begin{subfigure}[t]{0.13\columnwidth}
%    \centering\includegraphics[width=\textwidth]{04_cover_image_003.png}
%    \caption*{$I^{cov}$}
%\end{subfigure}
%\begin{subfigure}[t]{0.13\columnwidth}
%    \centering\includegraphics[width=\textwidth]{04_cover_image_003.png}
%    \caption*{$I^{cov}$}
%\end{subfigure}
%\begin{subfigure}[t]{0.13\columnwidth}
%    \centering\includegraphics[width=\textwidth]{05_hidden_image_003.png}
%    \caption*{$I^{con}$}
%\end{subfigure}
%\begin{subfigure}[t]{0.01\columnwidth}
%    \centering \hspace{1pt}
%\end{subfigure}
%\begin{subfigure}[t]{0.13\columnwidth}
%    \centering\includegraphics[width=\textwidth]{07_diff_cover_hidden_003.png}
%    \caption*{$I^{cov} - I^{con}$}
%\end{subfigure} \\

%% file: architecture_details.tex
\vspace{6pt}
\resizebox{400pt}{!}{%
\fontsize{4}{4}\selectfont
\centering
\small
    \begin{subtable}{.32\textwidth}
        \centering
        {\begin{tabular}{c | c | c }
        	\toprule
         		\multicolumn{3}{c}{\scriptsize $I^{sec} \in \mathbb{R}^{H \times W \times 1}$} \\
         		\midrule
        	    \scalebox{.9}{$Conv_{3 \times 3}^{50}$} & \scalebox{.9}{$Conv_{4 \times 4}^{50}$} & \scalebox{.9}{$Conv_{5 \times 5}^{50}$} \\
        	    \scalebox{.9}{$ReLU$} & \scalebox{.9}{$ReLU$} & \scalebox{.9}{$ReLU$} \\ \cmidrule(lr){1-1} \cmidrule(lr){2-2} \cmidrule(lr){3-3}
        	    \scalebox{.9}{$Conv_{3 \times 3}^{50}$} & \scalebox{.9}{$Conv_{4 \times 4}^{50}$} & \scalebox{.9}{$Conv_{5 \times 5}^{50}$} \\
        	    \scalebox{.9}{$ReLU$} & \scalebox{.9}{$ReLU$} & \scalebox{.9}{$ReLU$} \\ \cmidrule(lr){1-1} \cmidrule(lr){2-2} \cmidrule(lr){3-3}
        	    \scalebox{.9}{$Conv_{3 \times 3}^{50}$} & \scalebox{.9}{$Conv_{4 \times 4}^{50}$} & \scalebox{.9}{$Conv_{5 \times 5}^{50}$} \\
        	    \scalebox{.9}{$ReLU$} & \scalebox{.9}{$ReLU$} & \scalebox{.9}{$ReLU$} \\ \cmidrule(lr){1-1} \cmidrule(lr){2-2} \cmidrule(lr){3-3}
        	    \scalebox{.9}{$Conv_{3 \times 3}^{50}$} & \scalebox{.9}{$Conv_{4 \times 4}^{50}$} & \scalebox{.9}{$Conv_{5 \times 5}^{50}$} \\
        	    \scalebox{.9}{$ReLU$} & \scalebox{.9}{$ReLU$} & \scalebox{.9}{$ReLU$} \\
        	    \midrule
        	    \multicolumn{3}{c}{\scriptsize $Cat_{H \times W}^{150}$} \\
        	    \midrule
        	    \scalebox{.9}{$Conv_{3 \times 3}^{50}$} & \scalebox{.9}{$Conv_{4 \times 4}^{50}$} & \scalebox{.9}{$Conv_{5 \times 5}^{50}$} \\
        	    \scalebox{.9}{$Sigmoid$} & \scalebox{.9}{$ReLU$} & \scalebox{.9}{$Sigmoid$} \\ \cmidrule(lr){2-2}
        	    & \scalebox{.9}{$Conv_{4 \times 4}^{1}$} & \\
        	    & \scalebox{.9}{$Sigmoid$} & \\
        	    \midrule
        	    \multicolumn{3}{c}{\scriptsize $Cat_{H \times W}^{3}$} \\
        	    \midrule
        	    \multicolumn{3}{c}{\scriptsize $I^{pre} \in \mathbb{R}^{H \times W \times 3}$} \\
        	    \bottomrule
        \end{tabular}}
        \vspace{8pt}
        \caption*{\large \label{tab:preparation_net} Preparation Net $P_\theta$}
    \end{subtable}
	
    \begin{subtable}{.32\textwidth}
        \centering
        {\begin{tabular}{c | c | c}
          \toprule
          \multicolumn{3}{c}{\scriptsize $I^{pre}, I^{cov} \in \mathbb{R}^{H \times W \times 3}$} \\
          \midrule
          \multicolumn{3}{c}{\scriptsize $Cat_{H \times W}^{6}$} \\
          \midrule
          \scalebox{.9}{$Conv_{3 \times 3}^{50}$} & \scalebox{.9}{$Conv_{4 \times 4}^{50}$} & \scalebox{.9}{$Conv_{5 \times 5}^{50}$} \\
          \scalebox{.9}{$ReLU$} & \scalebox{.9}{$ReLU$} & \scalebox{.9}{$ReLU$} \\ \cmidrule(lr){1-1} \cmidrule(lr){2-2} \cmidrule(lr){3-3} 
          \scalebox{.9}{$Conv_{3 \times 3}^{50}$} & \scalebox{.9}{$Conv_{4 \times 4}^{50}$} & \scalebox{.9}{$Conv_{5 \times 5}^{50}$} \\
          \scalebox{.9}{$ReLU$} & \scalebox{.9}{$ReLU$} & \scalebox{.9}{$ReLU$} \\ \cmidrule(lr){1-1} \cmidrule(lr){2-2} \cmidrule(lr){3-3} 
          \scalebox{.9}{$Conv_{3 \times 3}^{50}$} & \scalebox{.9}{$Conv_{4 \times 4}^{50}$} & \scalebox{.9}{$Conv_{5 \times 5}^{50}$} \\
          \scalebox{.9}{$ReLU$} & \scalebox{.9}{$ReLU$} & \scalebox{.9}{$ReLU$} \\ \cmidrule(lr){1-1} \cmidrule(lr){2-2} \cmidrule(lr){3-3} 
          \scalebox{.9}{$Conv_{3 \times 3}^{50}$} & \scalebox{.9}{$Conv_{4 \times 4}^{50}$} & \scalebox{.9}{$Conv_{5 \times 5}^{50}$} \\
          \scalebox{.9}{$ReLU$} & \scalebox{.9}{$ReLU$} & \scalebox{.9}{$ReLU$} \\
          \midrule
          \multicolumn{3}{c}{\scriptsize $Cat_{H \times W}^{150}$} \\
          \midrule
          \scalebox{.9}{$Conv_{3 \times 3}^{50}$} & \scalebox{.9}{$Conv_{4 \times 4}^{50}$} & \scalebox{.9}{$Conv_{5 \times 5}^{50}$} \\
          \scalebox{.9}{$ReLU$} & \scalebox{.9}{$ReLU$} & \scalebox{.9}{$ReLU$} \\ \cmidrule(lr){1-1} \cmidrule(lr){2-2} \cmidrule(lr){3-3} 
          \scalebox{.9}{$Conv_{3 \times 3}^{1}$} & \scalebox{.9}{$Conv_{4 \times 4}^{1}$} & \scalebox{.9}{$Conv_{5 \times 5}^{1}$} \\
          \scalebox{.9}{$ReLU$} & \scalebox{.9}{$ReLU$} & \scalebox{.9}{$ReLU$} \\
          \midrule
          \multicolumn{3}{c}{\scriptsize $Conv_{1 \times 1}^{3}(Cat_{H \times W}^{150})$} \\
          \midrule
          \multicolumn{3}{c}{\scriptsize $I^{con} \in \mathbb{R}^{H \times W \times 3}$} \\
          \bottomrule
        \end{tabular}}
        \vspace{8pt}
        \caption*{\large \label{tab:hiding_net} Hiding Net $H_\phi$}
    \end{subtable}

    \begin{subtable}{.32\textwidth}
      \centering
      {\begin{tabular}{c | c | c}
          \toprule
          \multicolumn{3}{c}{\scriptsize $I^{con} \in \mathbb{R}^{H \times W \times 3}$} \\
          \midrule
          \scalebox{.9}{$Conv_{3 \times 3}^{50}$} & \scalebox{.9}{$Conv_{4 \times 4}^{50}$} & \scalebox{.9}{$Conv_{5 \times 5}^{50}$} \\
          \scalebox{.9}{$ReLU$} & \scalebox{.9}{$ReLU$} & \scalebox{.9}{$ReLU$} \\ \cmidrule(lr){1-1} \cmidrule(lr){2-2} \cmidrule(lr){3-3} 
          \scalebox{.9}{$Conv_{3 \times 3}^{50}$} & \scalebox{.9}{$Conv_{4 \times 4}^{50}$} & \scalebox{.9}{$Conv_{5 \times 5}^{50}$} \\
          \scalebox{.9}{$ReLU$} & \scalebox{.9}{$ReLU$} & \scalebox{.9}{$ReLU$} \\ \cmidrule(lr){1-1} \cmidrule(lr){2-2} \cmidrule(lr){3-3} 
          \scalebox{.9}{$Conv_{3 \times 3}^{50}$} & \scalebox{.9}{$Conv_{4 \times 4}^{50}$} & \scalebox{.9}{$Conv_{5 \times 5}^{50}$} \\
          \scalebox{.9}{$ReLU$} & \scalebox{.9}{$ReLU$} & \scalebox{.9}{$ReLU$} \\ \cmidrule(lr){1-1} \cmidrule(lr){2-2} \cmidrule(lr){3-3} 
          \scalebox{.9}{$Conv_{3 \times 3}^{50}$} & \scalebox{.9}{$Conv_{4 \times 4}^{50}$} & \scalebox{.9}{$Conv_{5 \times 5}^{50}$} \\
          \scalebox{.9}{$ReLU$} & \scalebox{.9}{$ReLU$} & \scalebox{.9}{$ReLU$} \\
          \midrule
          \multicolumn{3}{c}{\scriptsize $Cat_{H \times W}^{150}$} \\
          \midrule
          \scalebox{.9}{$Conv_{3 \times 3}^{50}$} & \scalebox{.9}{$Conv_{4 \times 4}^{50}$} & \scalebox{.9}{$Conv_{5 \times 5}^{50}$} \\
          \scalebox{.9}{$ReLU$} & \scalebox{.9}{$ReLU$} & \scalebox{.9}{$ReLU$} \\ \cmidrule(lr){2-2}
          & \scalebox{.9}{$Conv_{4 \times 4}^{1}$} & \\
          & \scalebox{.9}{$ReLU$} & \\
          \midrule
          \multicolumn{3}{c}{\scriptsize $Conv_{2 \times 2}^{1}(Cat_{H \times W}^{3})$} \\
          \midrule
          \multicolumn{3}{c}{\scriptsize $I^{rev} \in \mathbb{R}^{H \times W \times 1}$} \\
          \bottomrule
      \end{tabular}}
      \vspace{8pt}
      \caption*{\large \label{tab:reveal_net} Reveal Net $R_\gamma$}
    \end{subtable}
}